\def\year{2021}\relax
\documentclass[letterpaper]{article} 
\usepackage{aaai21}  
\usepackage{times}  
\usepackage{helvet} 
\usepackage{courier}  
\usepackage{amsfonts}
\usepackage{amsmath}
\usepackage{amssymb}
\usepackage{tabularx}
\usepackage{subfig}
\usepackage[hyphens]{url}  
\usepackage{graphicx} 
\usepackage{natbib}  
\usepackage{caption} 
\usepackage[english]{babel}
\usepackage{multirow}
\usepackage[T1]{fontenc}
\usepackage[utf8]{inputenc}

\usepackage{booktabs}
\title{An Empirical Comparison of Deep Learning Models for Knowledge Tracing on Large-Scale Dataset }
\author{
    
    Shalini Pandey,  
    George Karypis,
    Jaideep Srivastava
    \\
}
\affiliations{
    \textsuperscript{\rm 1} Department of Computer Science and Engineering\\University of Minnesota\\ Twin Cities, Minnesota, USA\\

}

\begin{document}

\maketitle

\begin{abstract}
Knowledge tracing (KT) is the problem of modeling each student's mastery of knowledge concepts (KCs) as (s)he engages with a sequence of learning activities. It is an active research area to help provide learners with personalized feedback and materials.Various deep learning techniques have been proposed for solving KT. Recent release of large-scale student performance dataset ~\cite{choi2019ednet} motivates the analysis of performance of deep learning approaches that have been proposed to solve KT. Our analysis can help understand which method to adopt when large dataset related to student performance is available. We also show that incorporating contextual information such as relation between exercises and student forget behavior further improves the performance of deep learning models.   
\end{abstract}

\section{Introduction}
The availability of large-scale student performance dataset has attracted researchers to develop models for predicting students' knowledge state aimed at providing proper feedback~\cite{self1990theoretical}. For developing such models, knowledge tracing (KT) is considered to be an important problem and is defined as tracing a student's \textit{knowledge state}, which represents her mastery level of KCs, based on her past learning activities. KT can be formalized as a supervised sequence learning task - given student's past exercise interactions \( \mathbf{X} = (\mathbf{x}_1, \mathbf{x}_2, \ldots, \mathbf{x}_t) \), predict some aspect of her next interaction $\mathbf{x}_{t+1}$. On the question-answering platform, the interactions are represented as
$\mathbf{x}_t = (e_t, r_t)$, where \( e_t \) is the exercise that the student attempts at timestamp $t$ and $r_t$ is the correctness of the student's answer. KT aims to predict whether the student will be able to answer the next exercise correctly, i.e., predict \( p(r_{t+1}=1| e_{t+1}, \mathbf{X}) \).\par
Among various deep learning models,  Deep Knowledge Tracing (DKT)~\cite{piech2015deep} and its variant~\cite{yeung2018addressing} use Recurrent Neural Network (RNN) to model a student's knowledge state in one summarized hidden vector. 
Dynamic Key-value memory network (DKVMN)~\cite{zhang2017dynamic} exploits Memory Augmented Neural Network~\cite{santoro2016one} for KT. It maps the exercises to the underlying KCs and then utilizes the student mastery at those KCs to predict whether the student will be able to answer the exercise correctly. The student mastery at the KCs are modeled using a dynamic matrix called \textit{value}. The student performance is then used to update the value matrix, thus updating student mastery at the associated KCs. Self-Attention model for Knowledge Tracing (SAKT) ~\cite{pandey2019self} employs a self-attention layer that directly identifies student past interactions that are relevant to the next exercise. It then predicts whether student will be able to solve the next exercise based on his/her performance at those past interactions. The assigned weights to the past interactions also provide an interpretation regarding which interactions from the past played important role in the prediction. Relation-Aware Self-Attention for Knowledge Tracing ~\cite{pandey2020rkt} improves over SAKT by incorporating contextual information and student forget behavior. It explicitly models  the relation between different exercise pairs from their textual content and student performance data and employs a kernel function with a decaying curve with respect to time to model student tendency to forget.   \par
\begin{figure*}[!ht]

     \subfloat[ DKT ]{%
      \includegraphics[width=0.21\textwidth]{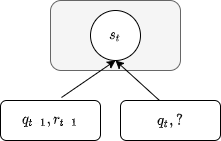}
    }
     \hfill
     \subfloat[ DKVMN ]{%
      \includegraphics[width=0.21\textwidth]{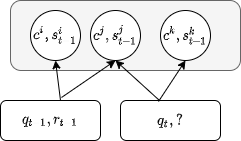}
     }
     \hfill
      \subfloat[ SAKT ]{%
      \includegraphics[width=0.26\textwidth]{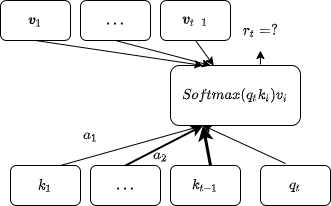}
     }
     \hfill
      \subfloat[RKT]{%
      \includegraphics[width=0.26\textwidth]{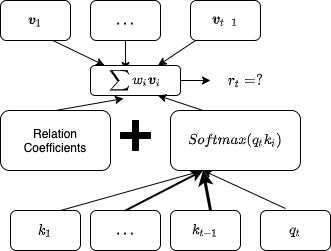}
     }
     \caption{Model differences among DKT, DKVMN, SAKT  and RKT. }
    \label{into}
  \end{figure*}
In this paper, we perform an analysis of the described deep-learning models for knowledge tracing. This analysis will help understand which deep-learning model performs best when we have massive student performance dataset. In addition, we visualize the attention weights to qualitatively reveal SAKT and RKT behavior.

To summarize, figure ~\ref{into} represents the difference between the four models we have analyzed in this work. First DKT uses a summarized hidden vector to model the knowledge state. Second, DKVMN maintains the concept state for each concept simultaneously
and all concept states constitute the knowledge state of a
student. Third, SAKT assigns weights to the past interaction using self-attention mechanism to identify the relevant ones. It then uses the weighted combination of these past interactions to estimate student knowledge on the involved KCs and predict her performance. Finally, RKT improves over SAKT by introducing a relation coefficient added to the attention weights learned from SAKT. The relation coefficient are learned from the contextual information explicitly modelling the relation between exercises involved in the past interactions and student forget behavior of students.

\section{Methods}
KT predicts whether a student will be able to answer the next exercise $e_{t}$ based on his/her previous interaction sequence ${X}={{x}_1, {x}_2, \ldots, {x}_{t-1}}$.  The deep learning methods transform the problem of KT into a sequential modeling problem. It is convenient to consider the model with inputs ${{x}_1, {x}_2, \ldots, {x}_{t-1}}$ and the exercise sequence with one position ahead, ${e_2, e_3, \ldots, e_t}$ and the output being the correctness of the response to exercises ${r_2, r_3, \ldots, r_t}$. 
 The interaction tuple \( {x}_t = (e_t,r_t) \) is presented to the model as a number $y_t = e_t+ r_t\times E$, where $E$ is the total number of exercises. Thus, the total values that an element in the interaction sequence can take is $2E$, while elements in the exercise sequence can take $E$ possible values.

\subsection{Deep Knowledge Tracing}
Deep Knowledge Tracing (DKT) ~\cite{piech2015deep} employs a Recurrent Neural Network (RNN) as its backbone model. 
As each interaction can be identified by a unique ID, it can be represented using the encoding vector $\textbf{x}_t$. After the transformation, DKT passes the 
$\textbf{x}_t$ to the hidden layer
and computes the hidden state $\textbf{h}_t$ using the vanilla RNN or
LSTM-RNN. As the hidden state summarizes the information
from the past, the hidden state in the DKT can be
treated as the latent knowledge state of student resulting
from  past learning trajectory. This latent knowledge state
is then fed to the output layer to compute the output
vector $\textbf{y}_{t}$, which represents the probabilities of answering each
question correctly. 
The objective of the DKT is to predict the next interaction
performance, so the target prediction is extracted by
performing a dot product of the output vector $\boldsymbol{y}_{t}$
and the one-hot encoded vector of the next question $\boldsymbol{e}_{t}$. Based on the
predicted output $p_{t}$ and the target output $r_{t}$, the loss function
$\mathcal{L}$ can be expressed as follows:
 \begin{equation}
     \mathcal{L}  = -\sum_t r_t \log(p_t) + (1-r_t) \log(1-p_t)
 \end{equation}

\subsection{Dynamic Key-Value Memory Network} 
Dynamic Key-Value Memory Network (DKVMN) ~\cite{zhang2017dynamic} exploits the relationship between concepts and the ability to trace each concept state. DKVMN model maps each exercise with the underlying concepts and maintains a concept state for each concept. At each timestamp, the knowledge of the related concept states of the attempted exercise gets updated. DKVMN 
consists two matrices, a static matrix called key, which stores
the concept representations and the other dynamic matrix
called value, which stores and updates the student’s understanding (concept state) of each concept.  
The memory, denoted as $M^t$, is an $N \times d$ matrix, where
$N$ is the number of memory locations, and $d$ is the embedding size. At each timestamp $t$, the input  is $\boldsymbol{x}_t$. The embedding vector $\boldsymbol{x}_t$ is used to compute the
read weight $\boldsymbol{w}^r_t$ and the write weight $\boldsymbol{w}^w_t$.
The intuition of
the model is that when a student answers the exercise that
has been stored in the memory with the same response, $\boldsymbol{x}_t$
will be written to the previously used memory locations and
when a new exercise arrives or the student gets a different response, $\boldsymbol{x}_t$ will be written to the least recently used memory
locations.
\par 
In DKVMN, when a student attempts an exercise, the student mastery over associated concepts is retrieved as weighted sum of all memory slots in the value matrix, where the weight is computed by taking the softmax activation of the inner
product between $\boldsymbol{x}_t$ and each key slot $\boldsymbol{M}^k(i)$.:
\begin{equation*}
    \boldsymbol{r}_t = \sum_{i=1}^{N} w_t(i) \boldsymbol{M}^v_t (i),
\end{equation*}
\begin{equation}
    w_t(i) =  \text{Softmax}(\boldsymbol{x}_t^T \boldsymbol{M}^k(i)).
\end{equation}
The calculated read content $\boldsymbol{r}_t$  is treated as a summary of the student’s mastery level of this exercise. Finally to predict the performance of the
student:
\begin{equation}
f_t = \text{Tanh}(\boldsymbol{W}_1^T [\boldsymbol{r}_t, \boldsymbol{x}_t] + \boldsymbol{b}_1),\\
    p_t = \text{Sigmoid}(\boldsymbol{W}_2^T \boldsymbol{f}_t + \boldsymbol{b}_2)
\end{equation}
\noindent $p_t$ is a scalar that
represents the probability of answering $e_t$ correctly. \par 
After the student answers the exercise $e_t$, the model 
updates the value matrix according to the correctness of the
student’s answer. For this, it  computes an erase vector $\boldsymbol{e}_t$ and an add vector $\boldsymbol{a}_t$ as:
\begin{equation}
    \boldsymbol{e}_t = \text{Sigmoid}(\boldsymbol{E}^T \boldsymbol{v}_t + \boldsymbol{b}_e), \boldsymbol{a}_t = \text{Tanh}(\boldsymbol{D}^T \boldsymbol{v}_t + \boldsymbol{b}_a)
\end{equation}

\noindent where the transformation matrices $\boldsymbol{E}, \boldsymbol{D} \in  \mathbb{R}^{d_v \times d_v}$. 

The memory vectors of value component $\boldsymbol{M}^v_{t}(i)$
from the previous timestamp are modified as follows:
\begin{align}
\boldsymbol{M}^v_t(i) = \boldsymbol{\Tilde{M}}^v_t (i) + w_t(i) \boldsymbol{a}_t,\\
    \boldsymbol{\Tilde{M}}^v_t (i) = \boldsymbol{M}^v_{t-1}(i)[\boldsymbol{1}- w_t(i)\boldsymbol{e}_t], 
\end{align}

\subsection{Self-Attention for Knowledge Tracing }
Self-Attention Model for Knowledge Tracing (SAKT) ~\cite{pandey2019self} is a purely transformer based model for KT. The idea behind SAKT is that in the KT task, the skills that a student builds while going through the sequence of learning activities, are related to each other and the performance on a particular exercise is dependent on his performance on the past exercises related to that exercise.  SAKT first identifies \textit{relevant} KCs from the past interactions and then predicts student's performance based on his/her performance on those KCs.  To identify the relevant interaction, it employs a self-attention mechanism which computes the dot-product between the past interaction representation and the next exercise representation. Essentially, the  student ability to answer next question is encoded in the vector $\textbf{y}_t$ and is computed as: 
\begin{equation}
    \textbf{y}_t = \sum_{j=1} ^ {t-1} \alpha_{j}{\textbf{x}}_j\textbf{W}^V, \alpha_j = \frac{\exp(e_{j})}{\sum_{k=1}^{t-1} \exp(e_{k})},\\ 
\end{equation}
\begin{equation}
   e_{j}=\frac{\textbf{e}_{t}\textbf{W}^Q({\textbf{x}}_j\textbf{W}^K)^T}{\sqrt{d}},
\end{equation}
\noindent where $d$ is the embedding size, $\textbf{W}^Q \in \mathbb{R}^{d\times  d}$, $\textbf{W}^V \in \mathbb{R}^{d\times d}$ and $\textbf{W}^K 
\in \mathbb{R}^{d\times d} $ are  projection matrices for query
and key, respectively.\\
\textit{Point-Wise Feed-Forward Layer:} In addition, a  PointWise Feed-Forward Layer (FFN) is applied to the output of SAKT.  The FFN helps incorporate non-linearity in the model and considers the interactions between different latent dimensions. It consists of two linear transformations with a ReLU nonlinear activation function between the linear transformations. The final output of FFN is $\textbf{F} =  \text{ReLU}(\textbf{y}_t \textbf{W}^{(1)} + \textbf{b}^{(1)}) \textbf{W}^{(2)}+\textbf{b}^{(2)}$, where $\textbf{W}^{(1)} \in \mathbb{R}^{d\times d}$, $\textbf{W}^{(2)} \in \mathbb{R}^{d\times d}$ are weight matrices and $\textbf{b}^{(1)} \in \mathbb{R}^{d}$ and $\textbf{b}^{(2)} \in \mathbb{R}^{d\times d}$ are the bias vectors. \par
Besides of the above modeling structure, we added residual connections ~\cite{he2016deep} after both self-attention layer and Feed forward layer to train a deeper network structure. We also applied the layer
normalization ~\cite{ba2016layer} and the dropout
~\cite{srivastava2014dropout} to the output of each layer, following ~\cite{vaswani2017attention}.
\subsection{Relation-aware Self-attention for Knowledge Tracing }
Similar to SAKT, Relation-aware Self-attention for Knowledge Tracing (RKT) ~\cite{pandey2020rkt} also identifies the past interaction relevant for solving the next exercise. Furthermore, it improves over SAKT  by incorporating contextual information. This contextual information integrates both the exercise relation information through their similarity as well as student performance data and the forget behavior information  through modeling an exponentially decaying  kernel function. Essentially, RKT exploits the fact that students acquire their skills while solving exercises and each such interaction has a distinct impact on student ability to solve a future exercise. This \textit{impact} is characterized by 1) the relation between exercises involved in the interactions and 2) student forget behavior.\par
RKT explicitly models the relation between exercises. To incorporate that it utilizes textual content of the exercises. In the absence of information about the textual content of exercises,we leaverage the skill tags associated with each exercise. Exercises $i,j$ with the same skill tag are given similarity value, $sim_{i,j}=1$, otherwise $sim_{i,j}=0$.
The correlation between exercises can also be determined from the learner's performance data . Essentially, RKT determines relevance of the knowledge gained from exercise $j$ to solve exercise $i$ by  building a contingency table as shown in table ~\ref{contingency}  considering only the pairs of $i$ and $j$, where $j$ occurs before $i$ in the learning sequence. Then it computes Phi coefficient that describes the relation from $j$ to $i$  as,
\begin{equation}
    \phi_{i,j} =\frac{n_{11}n_{00}-{n_{01}n_{10}}}{\sqrt{n_{1*}n_{0*}n_{*1}n_{*0}}} . 
\end{equation}
\begin{table}[]
\caption{A contingency table for two exercises $i$ and $j$.  }
\label{contingency}
\begin{tabular}{cccc|c}

\toprule
                            &           & \multicolumn{2}{c|}{exercise $i$} \\
                            
                            &           & incorrect      & correct   &total    \\
                            \midrule
\multirow{2}{*}{exercise $j$} & incorrect & $n_{00}$      & $n_{01}$ & $n_{0*}$    \\
                            & correct   & $n_{10}$      & $n_{11}$ & $n_{1*}$\\
                            \midrule
& total  & $n_{*0}$& $n_{*1}$& $n$\\
\bottomrule
\end{tabular}
\end{table}
Finally,  the relation of exercise $j$ with exercise $i$ is calculated as :
\begin{equation}
 \textbf{A}_{i,j}= 
\begin{cases}
    \phi_{i,j}+\text{sim}_{i,j} ,& \text{if } \text{sim}_{i,j}+\phi_{i,j}>\theta\\
    0,              & \text{otherwise},
\end{cases}
\end{equation}
where $\theta$ is a threshold that controls sparsity of relation matrix.  \par
RKT also models student forget behavior by employing a kernel function with exponentially decaying curve with time to reduce the importance of interaction as time interval increases following the idea from forgetting curve theory. Specifically, given the time sequence of interaction of a student $\textbf{t} = (t_1,t_2, \ldots ,t_{n-1}) $ and the time at which the student attempts next exercise $t_{n}$, we compute the relative time interval between the next interaction and the $i$th interaction as $\Delta_{i}=t_{n}-t_i$. Thus, we compute forget behavior based relation coefficients, $\textbf{R}^T = [\exp(-\Delta_{1}/S_u), \exp(-\Delta_{2}/S_u), \ldots, \exp(-\Delta_{n-1}/S_u)]$, where $S_u$  refers to relative strength of memory of student $u$ and is a trainable parameter in our model. The resultant relation coefficients 
\begin{equation}
    \textbf{R} = \text{softmax}(\textbf{R}^E + \textbf{R}^T),
\end{equation}

RKT also adopts the self-attention architecture ~\cite{vaswani2017attention}  similar to SAKT. To incorporate the relation coefficients into the learned attention weights, it adds the two weights: 
\begin{equation}
    \beta_{j}=\lambda \alpha_{j}+(1-\lambda)\textbf{R}_{j},
\end{equation}
where $\alpha$ is computed using Eq. 3. $\textbf{R}_j$ is the $j$th element of the relation coefficient $\textbf{R}$, $\lambda$ is a tunable parameter. The representation of output at the $i$th interaction, $\textbf{o}\in \mathbb{R}^d$, is obtained by the weighted sum of linearly
transformed interaction embedding and position embedding:
\begin{equation}
    \textbf{y}_t = \sum_{j=1} ^ {n-1} \beta_{j}\hat{\textbf{x}}_j\textbf{W}^V,
\end{equation}
\noindent where $\textbf{W}^V\in \mathbb{R}^{d\times d}$ is the projection matrix for value space.
The  further architecture  remains same as the SAKT described above. 
\begin{figure*}[h]
    \centering
     \subfloat[ AUC ]{%
      \includegraphics[width=0.48\textwidth]{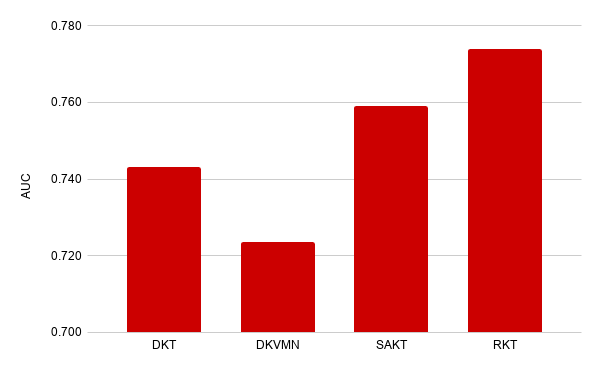}
    }
     \hfill
     \subfloat[ ACC]{%
      \includegraphics[width=0.48\textwidth]{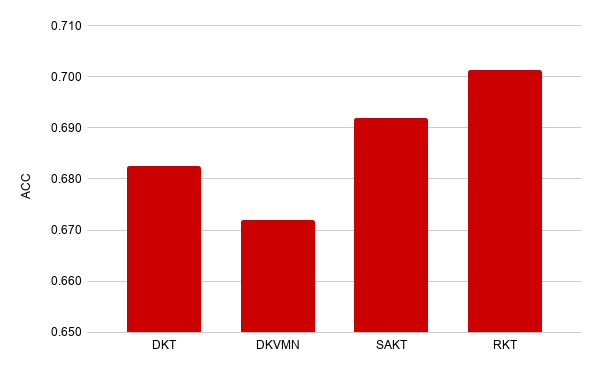}
     }
     
     \caption{Performance Comparison. RKT performs best among the models.  }
     \label{performance}
\end{figure*}


\section{Data}
To compare the deep-learning methods for KT, we use large-scale student interaction dataset, EdNet released in ~\cite{choi2019ednet}. EdNet consists of all student-system interactions collected over a
period spanning two years by Santa, a multi-platform AI tutoring service with
approximately 780,000 students. It has collected a total of 131,441,538 student interactions with each student generating an average of 441.20 interactions. The dataser consists a total 13,169 problems and 1,021 lectures tagged with 293 types of skills, and each of them has been consumed 95,294,926 times and
601,805 times, respectively. 
\section{Evaluation Setting}
The prediction of student performance is considered in a binary classification setting i.e., answering an exercise correctly or not. Hence, we compare the performance using the Area Under Curve (AUC) and Accuracy (ACC)  metric. 
Similar to evaluation procedure employed in ~\cite{nagatani2019augmenting, piech2015deep}, we train the model with the interactions in the training phase and during the testing phase, we update the model after each exercise response is received. The updated model is then used to perform the prediction on the next exercise.  Generally, the value $0.5$ of
AUC or ACC represents the performance prediction result
by randomly guessing, and the larger, the better. \par 
To ensure fair comparison, all models are
trained with embeddings of size $200$. The maximum allowed sequence length for self-attention is set as $50$. The model is trained with mini-batch size of $128$. We use Adam optimizer with a learning rate of $0.001$.
The dropout rate is set to $0.1$ to reduce overfitting. The L2 weight decay is set to $0.00001$. \par

\section{Results and Discussions}
\subsection{Quantitative Results}
Figure ~\ref{performance} shows the performance comparison of deep-learning models for KT on Ednet dataset. Different kinds of baselines demonstrate noticeable performance gaps. SAKT model shows improvement over DKT and DKVMN model which can be traced to the fact that SAKT identifies the relevance between past interactions and next exercise.  RKT performs consistently better than all the
baselines. Compared with other baselines, RKT is able to explicitly captures the relations between exercises based on student performance data and text content. Additionally, it models learner forget behavior using a kernel function which is more interpretable and proven way to model human memory ~\cite{ebbinghaus2013memory}.\par
The results reveal that  provided enough data, attention-based models surpass the other sequence encoder techniques such as RNN, LSTM and  Memory Augmented Networks. Furthermore, incorporating contextual data such as relation between exercises and domain knowledge such as student forget behavior attribute to performance gain even after availability of  the massive dataset. This motivates us to  further explore Knowledge Guided Machine Learning in the KT task. 
\begin{figure*}[!ht]

     \subfloat[ SAKT ]{%
      \includegraphics[width=0.48\textwidth]{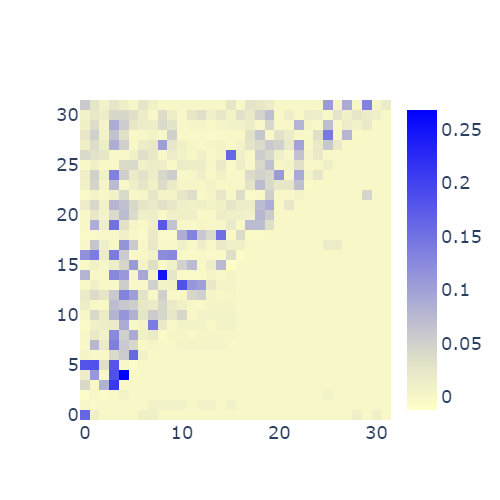}
    }
     \hfill
     \subfloat[ RKT ]{%
      \includegraphics[width=0.48\textwidth]{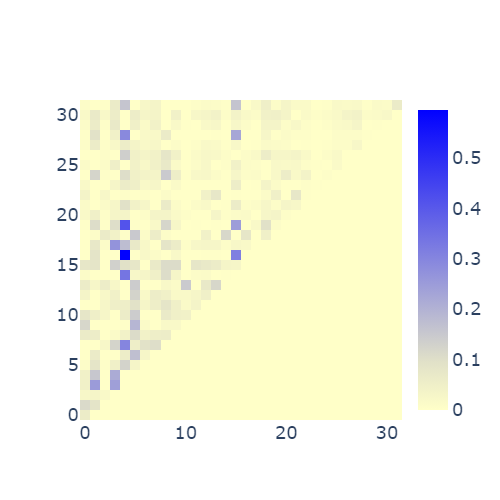}
     }
    
     \caption{Visualization of attention weights of an example student from EdNet by SAKT and RKT. Each subfloat depicts the attention weights assigned by the models for that student. }
    \label{heatmaps}
  \end{figure*}
  
  \begin{figure*}[!ht]

     \subfloat[ SAKT \label{subfloat-1}]{%
      \includegraphics[width=0.48\textwidth]{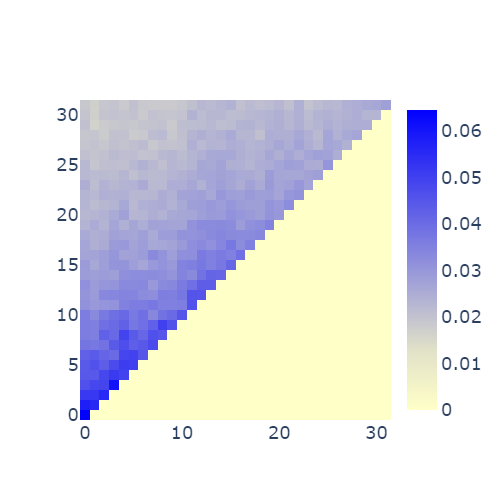}
    }
     \hfill
     \subfloat[ RKT \label{subfloat-2}]{%
      \includegraphics[width=0.48\textwidth]{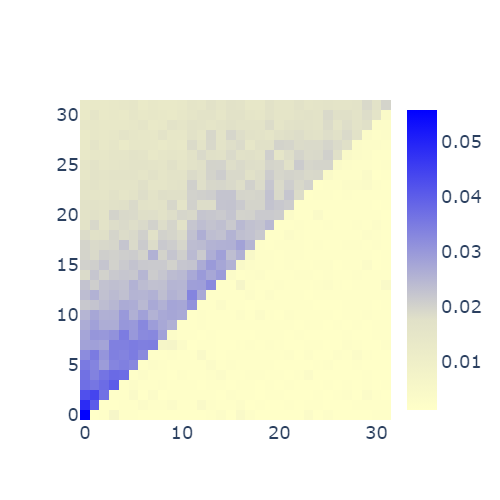}
     }
    
     \caption{Visualization of attention weights  pattern on different datasets. Each subfloat depicts the average attention weights of different sequences. }
    \label{vis}
  \end{figure*}
 
\subsection{Qualitative Analysis}
Benefiting from a purely attention mechanism, RKT and SAKT models are highly interpretable for explaining the prediction result. Such interpetability can help understand which past interactions played an important role in predicting student performance on the next exercise. To this end, we compared the attention weights obtained from both RKT and SAKT. We selected one student from the dataset and obtain the attention weights corresponding to the past interactions for predicting her performance at an exercise.   Figure ~\ref{heatmaps} shows the heatmap of attention weight matrix where $(i,j)$th element represents the attention weight on $j$th element when predicting performance on $i$th interaction.  We compare the generated heatmap  for both SAKT and RKT. This comparison shows the effect of relation information for revising the attention weights. Without relation information the attention weights are more distributed over previous interaction, while the relation information concentrates the attention weights to specific relevant interactions.\par

Finally we  also performed experiment to visualize the attention weights averaged over multiple sequences by RKT and SAKT. Recall that at time step $t_i$, the relation-aware self-attention layer in our model revises the attention weights on the previous interactions depending on the time elapsed since the interaction, and the relations between the exercises involved. To this end, we examine all sequences and seek to reveal meaningful patterns by showing the average attention weights on the previous interactions. Note that when we calculate the average weight, the denominator is the number of valid weights, so as to avoid
the influence of padding for short sequences. Figure ~\ref{vis} compares average attention weights assigned by SAKT and RKT. This comparison shows the effect of relation information for revising the attention weights. Without relation information the attention weights are more distributed over previous interaction, while the relation information concentrates the attention weights closer to diagonal.Thus, it is beneficial to consider relations between exercises for KT.
\section{Conclusion}
In this work, we analyzed the performance of various deep learning models for Knowledge Tracing. Analysis of these models on large dataset with approximately $780,000$ students revealed that self-attention based models such as SAKT and RKT outperform RNN-based models such as DKT. In addition, RKT which leverages additional information such as relation between exercises and student forget behavior and explicitly models these components gains further improvement. 
\bibliography{main.bib}

\begin{thebibliography}{13}
\providecommand{\natexlab}[1]{#1}
\providecommand{\url}[1]{\texttt{#1}}
\providecommand{\urlprefix}{URL }
\expandafter\ifx\csname urlstyle\endcsname\relax
  \providecommand{\doi}[1]{doi:\discretionary{}{}{}#1}\else
  \providecommand{\doi}{doi:\discretionary{}{}{}\begingroup
  \urlstyle{rm}\Url}\fi

\bibitem[{Ba, Kiros, and Hinton(2016)}]{ba2016layer}
Ba, J.~L.; Kiros, J.~R.; and Hinton, G.~E. 2016.
\newblock Layer normalization.
\newblock \emph{arXiv preprint arXiv:1607.06450} .

\bibitem[{Choi et~al.(2019)Choi, Lee, Shin, Cho, Park, Lee, Baek, Kim, and
  Jang}]{choi2019ednet}
Choi, Y.; Lee, Y.; Shin, D.; Cho, J.; Park, S.; Lee, S.; Baek, J.; Kim, B.; and
  Jang, Y. 2019.
\newblock EdNet: A Large-Scale Hierarchical Dataset in Education.
\newblock \emph{arXiv} arXiv--1912.

\bibitem[{Ebbinghaus(2013)}]{ebbinghaus2013memory}
Ebbinghaus, H. 2013.
\newblock Memory: A contribution to experimental psychology.
\newblock \emph{Annals of neurosciences} 20(4): 155.

\bibitem[{He et~al.(2016)He, Zhang, Ren, and Sun}]{he2016deep}
He, K.; Zhang, X.; Ren, S.; and Sun, J. 2016.
\newblock Deep residual learning for image recognition.
\newblock In \emph{Proceedings of the IEEE conference on computer vision and
  pattern recognition}, 770--778.

\bibitem[{Pandey and Karypis(2019)}]{pandey2019self}
Pandey, S.; and Karypis, G. 2019.
\newblock A Self-Attentive model for Knowledge Tracing.
\newblock \emph{arXiv preprint arXiv:1907.06837} .

\bibitem[{Pandey and Srivastava(2020)}]{pandey2020rkt}
Pandey, S.; and Srivastava, J. 2020.
\newblock RKT: Relation-Aware Self-Attention for Knowledge Tracing.
\newblock In \emph{Proceedings of the 29th ACM International Conference on
  Information \& Knowledge Management}, 1205--1214.

\bibitem[{Piech et~al.(2015)Piech, Bassen, Huang, Ganguli, Sahami, Guibas, and
  Sohl-Dickstein}]{piech2015deep}
Piech, C.; Bassen, J.; Huang, J.; Ganguli, S.; Sahami, M.; Guibas, L.~J.; and
  Sohl-Dickstein, J. 2015.
\newblock Deep knowledge tracing.
\newblock In \emph{Advances in Neural Information Processing Systems},
  505--513.

\bibitem[{Santoro et~al.(2016)Santoro, Bartunov, Botvinick, Wierstra, and
  Lillicrap}]{santoro2016one}
Santoro, A.; Bartunov, S.; Botvinick, M.; Wierstra, D.; and Lillicrap, T. 2016.
\newblock One-shot learning with memory-augmented neural networks.
\newblock \emph{arXiv preprint arXiv:1605.06065} .

\bibitem[{Self(1990)}]{self1990theoretical}
Self, J. 1990.
\newblock Theoretical foundations for intelligent tutoring systems.
\newblock \emph{Journal of Artificial Intelligence in Education} 1(4): 3--14.

\bibitem[{Srivastava et~al.(2014)Srivastava, Hinton, Krizhevsky, Sutskever, and
  Salakhutdinov}]{srivastava2014dropout}
Srivastava, N.; Hinton, G.; Krizhevsky, A.; Sutskever, I.; and Salakhutdinov,
  R. 2014.
\newblock Dropout: a simple way to prevent neural networks from overfitting.
\newblock \emph{The journal of machine learning research} 15(1): 1929--1958.

\bibitem[{Vaswani et~al.(2017)Vaswani, Shazeer, Parmar, Uszkoreit, Jones,
  Gomez, Kaiser, and Polosukhin}]{vaswani2017attention}
Vaswani, A.; Shazeer, N.; Parmar, N.; Uszkoreit, J.; Jones, L.; Gomez, A.~N.;
  Kaiser, {\L}.; and Polosukhin, I. 2017.
\newblock Attention is all you need.
\newblock In \emph{Advances in Neural Information Processing Systems},
  5998--6008.

\bibitem[{Yeung and Yeung(2018)}]{yeung2018addressing}
Yeung, C.-K.; and Yeung, D.-Y. 2018.
\newblock Addressing two problems in deep knowledge tracing via
  prediction-consistent regularization.
\newblock \emph{arXiv preprint arXiv:1806.02180} .

\bibitem[{Zhang et~al.(2017)Zhang, Shi, King, and Yeung}]{zhang2017dynamic}
Zhang, J.; Shi, X.; King, I.; and Yeung, D.-Y. 2017.
\newblock Dynamic key-value memory networks for knowledge tracing.
\newblock In \emph{Proceedings of the 26th International Conference on World
  Wide Web}, 765--774. International World Wide Web Conferences Steering
  Committee.

\end{thebibliography}
\end{document}